\documentclass[runningheads]{llncs}

\usepackage[english]{babel}
\usepackage{combelow}
\usepackage{wrapfig,lipsum,booktabs}
\usepackage{graphicx}
\usepackage{algorithm,algpseudocode}
\usepackage{amsmath}
\usepackage{amsfonts}
\usepackage{amssymb}

\title{Socially Inspired Communication in Swarm Robotics}

\author{Nathan White \and John Harwell \and Maria Gini}
\institute{Department of Computer Science and Engineering, University of Minnesota\\
\email{\{whit2267,harwe006,gini\}@umn.edu}}

\begin{document}

\maketitle

\begin{abstract}
  Localized communication in swarms has been shown to increase swarm effectiveness in
  some situations by allowing for additional opportunities for cooperation. However,
  communication and utilization of potentially outdated information is also a
  concern. We present an explicit non-directional goal-based communication model and
  message accept/reject scheme, and test our model in a set of object gathering
  experiments with a swarm of robots. The results of the experiments indicate that even low levels of communication
  regarding the swarm's goal outperform high levels of random information
  communication.


\end{abstract}

\section{INTRODUCTION}

Swarm robotics is the study of large scale robotic systems that consist of many
individual robots working cooperatively to achieve a goal
\cite{Sahin,Sharkey2007,Brutschy2012,Labella,Lerman,Harwell2018b}. These individual
robots usually have limited capabilities, so it is very difficult and time consuming for a single robot to achieve the goal. Cooperation among the robots helps the swarm to achieve the goal more robustly.  In a swarm each robot is autonomous, acting without a centralized controller. This
allows for heterogeneous adaptation to environmental differences in spatially disparate
parts of the swarm's operating area.

One popular application of swarm robotics is foraging. Foraging is the act of having
robots grab blocks, representing food, in a given environment and return them to a
central location (the ``nest''). Foraging in swarm robotics attempts to
replicate the efficiency observed in nature \cite{Mohan2009}.
This replication of cooperation found in nature has multiple real-world applications.
Companies such as NASA are considering swarms of robots that are able to cooperate
and work together for potential excursions into the asteroid belt
\cite{Rouff}. This would allow for cooperative exploration and communication
back to Earth. Another application is the usage of swarm robotics to complete tasks within
potentially dangerous regions \cite{Sahin}. One example of such tasks is the exploration of a
burning building. Swarm robotics would allow for the searching of people to be rescued, with
potentially lower search times and more flexibility, as the robots an search in parallel in different parts of the building.

Foraging tasks in swarm robotics has long been known to have better efficiency when
communication is permitted \cite{Balch2005}. This is due to the levels of cooperation
that can be achieved when sharing information via communication. Any robust communication model should be
able to increase swarm effectiveness. Furthermore, it should be able to
determine what information is relevant to increasing the effectiveness and what is
not, in order to minimize time lost due to out-of-date information.

In the realm of human communication, humans are able to communicate with any number
of people within a given range, where the range is limited only by hearing capabilities and the speaker's
volume. Humans are extremely good at cooperative work primarily due to their ability
to communicate \cite{Smith2010a}. Like in swarm robotics, humans are independent agents, 
acting according to their internal knowledge and representation of the
environment. However, unlike in swarm robotics, humans are capable of making
irrational decisions \cite{Lee2006a}, even forgoing given information if they believe
their internal representation is more accurate than the information that was passed to them by communication.

While ignoring information is an important part of the process, it is useless without
the ability to share information. Creating a communication model in swarm robotics
based on humans means the model has to include both the chance of ignoring information and the
chance of sharing information.

In this paper, we focus on a foraging scenario, where groups of robots have to gather
blocks from a single source and transport them to a known nest location. Robots need
to be capable of communication for improved cooperation opportunities. Within our
scenario, we allow explicit non-directional communication of source locations to
avoid wasted exploration time, but also allow for the potential to reject the
integration of a message. By utilizing communication, we can expect to see more
blocks being collected and less time spent in the exploration state.

We propose a new communication strategy for cooperative swarm robotics that utilizes
a form of explicit non-directional communication. We explore this strategy's
effectiveness within an ideal foraging scenario simulation, comparing it against
a random cell selection (RCS) algorithm with high levels of communication, as well as a controlled
random walk (CRW) swarm with no communication or memory of their environment. We view an ideal foraging
scenario as one without obstacles with goal objects located in a consistent location. The 
results of our experiments indicate that any level of the communication of information relevant to
the swarm's goal outperforms continuous information of random portions of the environment, but
any communication outperforms no communication.

The remainder of this paper is split into six sections. In Section
\ref{sec:relatedworks} we give a review of current applications and implementations
of communication in swarm robotics. Then, in Section
\ref{sec:problemstatementandproposedmethod} we provide an in-depth analysis of the
foraging scenario and solve it using our proposed communication
implementation. Section \ref{sec:experimentalframework} provides details for both the
framework and the assumptions we use in our experimental setup.  In
Section \ref{sec:experiments} and Section \ref{sec:results} we describe in detail the experiments,
followed by their results. The final section, Section \ref{sec:conclusion}, completes
the paper, with the conclusions and potential ideas for  future work.


\section{Related Work} \label{sec:relatedworks}
Communication strategies in swarm robotics are often inspired by ethology, the study
of animal behavior. This is  due to the fact the  in the animal kingdom many creatures are social and operate collectively to achieve their  goals. Several strategies have roots
in the studies of bees and ants \cite{Labella,Hecker2015,Ducatelle2011a}. This is
due to the fact that bees and ants  commonly represent the two main methods for communication,
explicit and implicit respectively.

Implicit communication is the use of the environment to share information with other
individuals. In the case of ants, pheromone trails are utilized to mark the path traversed. Pheromone trails  have
been replicated in prior swarm robotics research \cite{Labella,Hecker2015,Payton2003,Sumpter2003,Arab2012}.
Pheromone is left behind on the path an ant
takes. The pheromone decays over time, so repeated usage of the trails strengthens them. The stronger the level of
pheromone, the more ants are attracted to that specific pathway. In this way ants find the shortest paths. 

Conversely, explicit communication is the act of communicating directly with other
entities \cite{Trianni2004,Arvin2010a}. This can be done in many ways. In the
case of bees, the medium is a form of dance, known as the waggle dance
\cite{Frisch1967,Riley2005a}. This dance may need to be repeated if the bees fail to
find the location encoded within it. Using this method, robots have danced in
order to communicate source locations to the rest of the swarm
\cite{Pitonakova2016a}.

Regardless of the medium, the purpose is clear: to recruit other members of the swarm for
cooperative task completion. There have been many variants in the implementations, all to
increase the swarm effectiveness given their specific situation \cite{Mohan2009}. However, it
is clear that communication  is
useful to increase the swarm effectiveness in accomplishing the task.

Arkin \textit{et al.}, explored state based communication, where robots are only allowed to
communicate their current task, purely as an
aid, not as a necessary component in task completion \cite{Arkin1993a,Arkin1992a}. 
Utilizing a shared memory location, agents iteratively update their
current state and location. Communication is only utilized when a robot has no goals
in its field of view. If no goals are within view, then the robot is able to 
access the shared memory location to find which robots have found a goal and
where their location is, then is able to navigate in that direction.

However, while Arkin explored the usage of state-based communication, Balch
studied the effects of goal and state based communication over no
communication \cite{Balch2005}.  He noted that goal based
communication, the communication of locations of a goal object or place,
within a foraging scenario demonstrated a notable improvement over non
communicating swarms, but only a small improvement over state based communication. To
give our communication schema ideal conditions, we follow the principles of goal
based communication, being able to transmit source locations to others within the
swarm.

This has been explored further by Pugh
\cite{Pugh2013}. Entities are able traverse the environment in search of a single
food source. However, the food source requires three individual robots to lift it and
move it to a nest location. Pugh \textit{et al.} state that communication is promoted
by this need for several entities to lift and transport the food. By communicating,
the robots are able to gain more food through the course of the experiment, and spend
less time exploring.

Arkin and Pugh aren't alone in their studies. Many researchers have utilized
communication in order to increase their swarms effectiveness and ability to
cooperate (e.g.,\cite{Labella,Hecker2015,Payton2003,Sumpter2003,Arab2012,Pitonakova2016a,Pini2013c}). 
However, what is missing on all these studies is the ability of the robots
to reject communication. As given in our description of swarm robotics, robots are
individuals and as such can make decisions about their environment and the
information available to them. This should include the information shared with them.


\section{Problem Statement and Proposed Method} \label{sec:problemstatementandproposedmethod}
\subsection{Problem Statement} \label{subsec:problemstatment}

Each robot keeps a 2D grid of its environment. We denote a unit area of this 2D grid
as cell $(i,j)$. Each cell consists of two layers: the first being the contents
of $(i,j)$, which is represented by $s\in\{ Unknown, Empty, Has\_Block\}$, and the second
layer is the pheromone level associated with $(i,j)$.

When encountering a block within the environment, we say that a robot $k$ visited
cell $(i,j)$ at time step $t$. As time progresses, after $n$ time steps past $t$, in
which robot $k$ does not see the given cell, the pheromone will decay as in
Equation (\ref{eq:pheromone_decay}), where $\tau_{ij}$ represents the pheromone
level of cell $(i,j)$, $\tau_{ij}^k$ is robot $k$'s perception of pheromone levels at
that cell location, $\rho\in[0,1]$ is the pheromone decay parameter that
controls the rate of decay, for our experiments we set n = 1 (the level is updated at every time step),
and m = 1 (the amount of pheromone deposited per time step) \cite{Meng2008}.

\begin{equation} \label{eq:pheromone_decay}
    \tau_{ij}(t + n) = \rho\tau_{ij}^k(t) + \sum_{k=1}^m \Delta \tau_{ij}^k(t)
\end{equation}

The pheromone decay function dictates how relevant a cell's information is. If a
robot $k$ receives a message $m_i$ at time step $t$, denoted as $m_i(t)$, it should
have an associated relevance given by Eqn.~\eqref{eq:pheromone_decay}. Since the pheromone level
of cell $(i,j)$ indicates how relevant its information is. If the communicated pheromone
level is lower than the current internal level that robot $k$ has for that region, then
the communicated information is potentially outdated and would be rejected (e.g., if robot $k$ sends a message to robot $l$ where $\tau_{ij}^k < \tau_{ij}^l$ for
a cell $(i,j)$ then robot $l$ will reject the message).

Every robot is capable of sending at most 1 message per time step. Should robot $k$
send a message $m_i$, every robot within a radius of $r_k$ of robot $k$ will have the
message broadcast to them. A robot $k$ has a probability $p_{send}(t)$ and $p_{receive}(t)$ of sending
and receiving a message on a time step $t$, respectively. While probabilistic
message transmission is not new, probabilistic message reception is new, and models (1) potentially bad
environmental conditions that could cause unreliable communications, and (2) robot $k$'s uncertainty about
the trustworthiness of robot $l$'s information.

Under this problem definition, swarms collectively solve a multi-objective optimization problem: minimizing the number of inaccuracies within each robot's internal 
representation of block locations ($I(N)$) while simultaneously trying to 
maximize the total number of blocks gathered ($B(N)$).

\begin{equation}
\max{{B(N)}\min{I(N)}}
\end{equation}

Inaccuracies are calculated when cell $(i,j)$ enters robot $k$'s line of sight.
If the cell's actual state doesn't match the state of the robot's internal representation,
it is marked as inaccurate and recorded.

We therefore measure swarm performance in terms of this multi-objective formulation:

\begin{equation} \label{eq:swarm_performance}
P(N) = \frac{B(N)}{I(N)}
\end{equation}

\subsection{Proposed Method} \label{subsec:proposedmethod}
\begin{figure}[t!]
  \centering
  \includegraphics[keepaspectratio, width=0.65\textwidth]{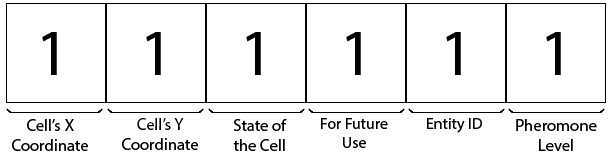}
  \caption{Communication data packet structure. Each packet is represented as a byte
    vector, meaning each piece of information can be represented with a maximum of 1
    byte. The 1 in each box represents the 1 byte limit, and each box is labeled
    with the information it is representing.}
  \label{fig:CommunicationStructure}
\end{figure}

Before discussing the algorithm, we introduce the communication packet structure.
Each packet is limited to 6 bytes of data, the structure for which is shown in Fig.
\ref{fig:CommunicationStructure}. The first two bytes represent the (X, Y)
coordinates of the cell $(i,j)$. The third byte refers to the sending robot's internal
knowledge of the current state of $(i,j)$, which in our constrained foraging
scenario is a subset of the complete set of states a cell can have. The set of cell
states that we are interested in can be formulated as $s\in\{ Unknown, Empty, Has\_Block\}$,
where $s$ is the current state of the cell $(i,j)$. The fourth byte is reserved for future
use. The fifth byte represents the ID of the entity located in cell $(i,j)$.
Finally, the sixth byte represents the pheromone
level of the sending robot for the cell $(i,j)$.

\begin{algorithm}
\begin{algorithmic}[1]
\Statex
\Function{Communicate}{}
    \State {$p_{send}$ $\gets$ {$rand()$}}
    \State {$p_{receive}$ $\gets$ {$rand()$}}
    \If{$p_{receive} < \beta_{receive}$}{}
            \State {$Integrate\_Messages()$}
    \EndIf

    \If{$p_{send} < \beta_{send}$}{}
            \State {$cell$ $\gets$ {$SELECT\_CELL()$}}
            \State {$send(cell)$}
    \EndIf
\EndFunction
\Function{Select\_Cell}{}
     \State {return $\max_{i,j}{\frac{num_{ij}}{dist_{ij}} * \tau_{ij}(t)}$} \label{eq:utility}
\EndFunction
\end{algorithmic}
\caption{$\beta_{send}$ and $\beta_{receive}$ are experimental parameters.} 
\label{alg:communication}
\end{algorithm}

We utilize the explicit (sometimes called direct) communication strategy. 
At each time step of the simulation, robot $k$
probabilistically sends one communication packet to every robot $l$ in radius $r_k$ defined
by probability $p_{send}$. Similarly, each robot $l$ in the radius probabilistically
receives messages at each time step, defined by $p_{receive}$.

When a message is received, if robot $k$ decides to accept message $m_i$, and internalize
its contents, it treats all communicated data as if it was its own. That is to say
that all communicated observations within the swarm are treated as if each individual
robot had made the observation, when robot $k$ accepts it. The process of internalizing
the packet contents involves accessing robot $k$'s 2D grid of the environment.

When robot $k$ decides to send a message, it utilizes Algorithm \ref{alg:communication} 
in order to select a cell $(i,j)$ that maximizes line~\eqref{eq:utility}. The criteria for
this equation are the number of blocks within the cell, $num_{ij}$, the euclidean distance from the 
cell to the nest, $dist{ij}$, and the pheromone level associated with the cell, $\tau_{ij}$.
Maximizing this function ensures trustworthy information is balanced
with valuable information by trying to maximize both $num_{ij}$ and $\tau_{ij}$ while minimizing
$dist_{ij}$. For example, in the event there is a large store of blocks
close to the nest with a low level of associated pheromone, it might be better to inform
nearby robots of a different location, even if said location contains fewer blocks and
lies just further away.


\section{Experimental Framework} \label{sec:experimentalframework}

To conduct the experiments mentioned in this paper, we utilized the open-source
FORDYCA \cite{Harwell2018b} project, built on the ARGoS \cite{Pinciroli2010}
simulator. The simulation's robots are modeled after an s-bot, developed during the
Swarm-bots project \cite{Dorigo2005}.

The results of each experiment is averaged over 50 simulations. 
For all experiments conducted, we make the following assumptions:
\begin{itemize}
    \item The robots are homogeneous, have an unlimited battery supply, and are able to communicate directly through range and bearing sensors.
    \item All robots perform the entire foraging task.
    \item Robots are randomly distributed in the environment, but are able to self
      localize based upon a known light source that resides above the nest.
    \item The arena size is known to the robots, but not its contents.
    \item Transfer of objects between robots is not permitted.
    \item All foraging takes place in a flat, obstacle-less environment.
    \item The capacity of the nest is not limited.
\end{itemize}

\begin{figure}[t]
  \centering
  \includegraphics[keepaspectratio, width=0.5\textwidth]{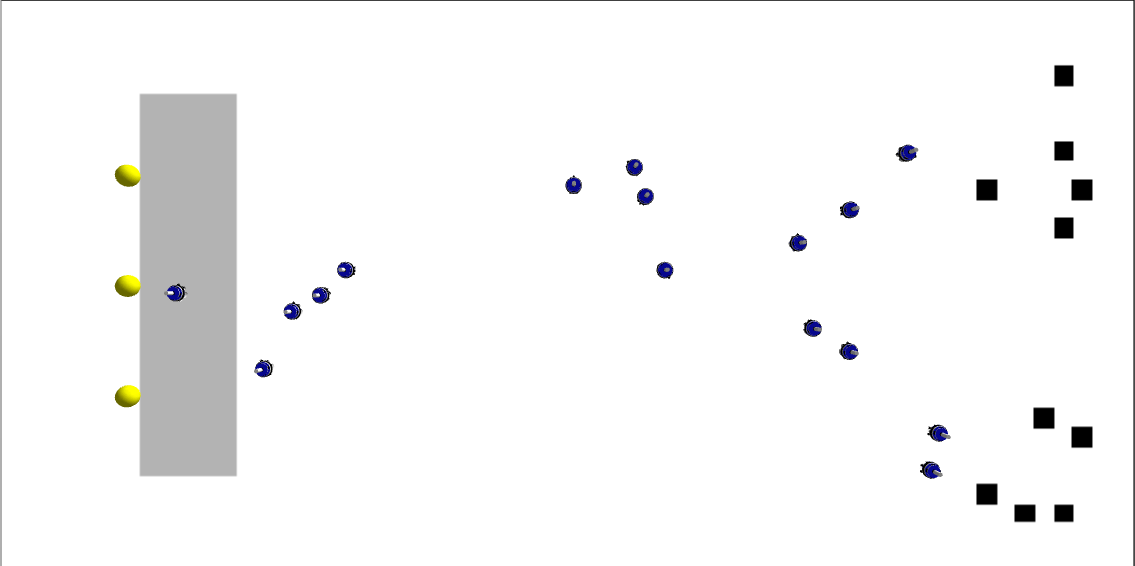}
  \caption{ARGoS foraging scenario where blocks, represented as black cubes, are
    clustered in a single source on the right end of the arena. The grey rectangle
    represents the nest, and is colored due to the robot's ground sensors only
    detecting gray-scale. The yellow spheres above the nest are the light sources the
    robots use for self localization.}
  \label{fig:Arena}
\end{figure}


\section{Experiments} \label{sec:experiments} 
We test our proposed method on nine
different sets of experiments in order to compare its performance against that of a
similar communication schema with the cell selection method as random (RCS) using high
probabilities for both sending and receiving. We also compare these results against a
swarm with no communication that explores its environment through random movement (CRW),
but retains no knowledge or assumptions about the location of food sources. Swarm performance
is measured by Eqn.~\eqref{eq:swarm_performance}.

\begin{table}[h]
 \centering
 \caption{Summary of parameters used for all experiments}
\label{table:parameters}
 \begin{tabular}{|c|c|}
 \hline
  Parameter & Value \\
 \hline
 $r_k$ &  2 \\
 \hline
  $\rho$ & 0.001 \\
 \hline
 Low & 30\% \\
 \hline
 Medium & 60\% \\
 \hline
 High & 90\% \\
 \hline
\end{tabular}
\end{table}

\begin{table}[h]
\centering
\caption{Summary of the experimental scenarios used for testing the proposed method.}
\label{table:experiments}
\begin{tabular}{|c|c|c|}
 \hline
  Experiment Set & $\beta_{send}$ & $\beta_{receive}$ \\
 \hline
 1 & Low & Low \\
 \hline
 2 & Low & Medium \\
 \hline
 3 & Low & High \\
 \hline
 4 & Medium & Low \\
 \hline
 5 & Medium & Medium \\
 \hline
 6 & Medium & High \\
 \hline
 7 & High & Low\\
 \hline
 8 & High & Medium \\
 \hline
 9 & High & High\\
 \hline
\end{tabular}
\end{table}

Table \ref{table:parameters} summarizes the values of the parameters that were kept constant
throughout the experiments. The value $r_k$ was selected to achieve a reliable
communication distance that remained realistic in an area proportional to the robot
size. That is to say, the area for communication potential is not excessively large
nor excessively small. The value chosen for $\rho$ strikes a good balance
between information relevance degradation and keeping viable blocks around long
enough to prevent premature lapse into irrelevance. Low, Medium, and High refer to the
probability for the sending/receiving probabilities, and are used in Table \ref{table:experiments}
to better convey the static associated value.

Table \ref{table:experiments} displays a summary of the experiments conducted. We
explore  varying the communication probabilities at several fixed probabilities to
determine where swarm effectiveness is maximized, while reducing the number of inaccuracies
in internal environment representation. All nine sets of experiments are
conducted with 128 robots, as well as a total of 75 source blocks located on the right
end of the arena. All experiments were conducted using the arena displayed in Figure
\ref{fig:Arena}.


\section{Results} \label{sec:results}
For each experiment, we measure the total number of blocks gathered at the end of the experiment as
well as the number of inaccuracies at every $\Delta t = 1000$ time steps. We define our swarm performance
$P(N)$ as being the total number of blocks collected divided by the number of inaccuracies recorded.

\begin{table}[h]
    \centering
        \caption{Results of Experiments 1-9, the random cell selection algorithm (RCS),
            and the controlled random walk (CRW) swarm,
            with the total number of blocks collected and the total number
            of inaccuracies averaged 50 simulations per experiment, as well as the
            swarm performance as defined in Eqn.~\eqref{eq:swarm_performance}.}
    \begin{tabular}{|c|c|c|c|}
        \hline
        Experiment & Average Blocks Collected & Average Inaccuracies & Swarm Performance \\
        \hline
        1 & 995.88 & 1469.956 & 0.6775 \\
        \hline
        2 & 991.26 & 1469.595 & 0.6745 \\
        \hline
        3 & 972.54 & 1470.362 & 0.6614 \\
        \hline
        4 & 975.94 & 1485.119 & 0.6571 \\
        \hline
        5 & 989.76 & 1468.137 & 0.6742 \\
        \hline
        6 & 1002 & 1447.042 & 0.6924 \\
        \hline
        7 & 979.82 & 1453.959 & 0.6739 \\
        \hline
        8 & 989.7 & 1475.9 & 0.6706 \\
        \hline
        9 & 995.22 & 1502.555 & 0.6624 \\
        \hline
        \hline
        RCS & 637.44 & 3554.704 & 0.1793 \\
        \hline
        CRW & 373.52 & 0 & NaN \\
        \hline
    \end{tabular}
    
    \label{tab:results}
\end{table}

\begin{figure}[h!]
  \centering
  \includegraphics[keepaspectratio, width=0.7\textwidth]{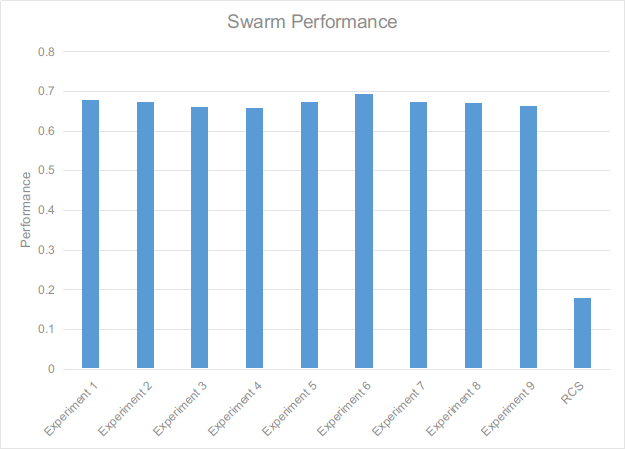}
  \caption{Visualization of the performance of the Experiments 1-9 and RCS. CRW not depicted.}
  \label{fig:Performance}
\end{figure}

The results indicate that even with a low chance of 
communication, information relevant to the goal of the swarm is much better than
always communicating potentially random information, but that any level of
communication outperforms swarms without it. RCS also had over double 
the number of inaccuracies regarding block locations than Experiment 9, the worst
performing experiment. The performance of each experiment and RCS can be observed
in Figure \ref{fig:Performance}, where the difference between RCS and utility
based selection becomes very apparent ($\sim47\%$). Due to CRW not retaining 
knowledge of its environment, it has zero inaccuracies, however is included in
our experiments to shows the performance difference of having any form of communication
versus having none.

The similarities between both the low communication in Experiment 1 and the high
communication in Experiment 9 indicate that communication occurs frequently enough that no
additional useful information is communicated at higher levels. More specifically, the cell that
was selected from the result of the utility function didn't vary frequently enough to warrant
excess communication.


\section{Conclusions and Future Work} \label{sec:conclusion}

We have presented a new communication schema for foraging in swarm robotics, adding
the ability for robots to reject messages, an ability not present in previous
work. We have shown that using this model, any level of relevant communication
outperforms constant communication of random information.

One possible direction for future work would involve the presence of dynamic task
allocation and caches. This would allow us to expand our communication
implementation and include a combination of state and goal based communication to
evaluate the impact it would have on task assignment and swarm efficiency. Another
avenue for further work would be the testing of this implementation in more dynamic
environments, where blocks are placed randomly or according to some function, as
opposed to in a single location. With both of these possibilities, we plan to
explore other communication algorithms and how the performance compares between
them and the one presented in this paper.

In an effort to facilitate collaboration and future research, the code for this work
is open source and available on github at
$https://github.com/swarm-robotics/fordyca.git$.

\subsubsection*{Acknowledgements.} We gratefully acknowledge Amazon Robotics, the MnDRIVE RSAM initiative at the
University of Minnesota, and the
Minnesota Supercomputing Institute (MSI) for their support of this work.

\bibliographystyle{splncs04}
\bibliography{refs}

\end{document}